\documentclass[11pt]{article}
\usepackage{acl}
\usepackage{times}
\usepackage{latexsym}
\usepackage{hyperref}

\usepackage[T1]{fontenc}

\usepackage[utf8]{inputenc}

\usepackage{microtype}

\usepackage{inconsolata}

\usepackage{graphicx}

\usepackage{graphicx}
\usepackage{soul}
\usepackage{bm} 
\usepackage{amsmath} 
\usepackage{amssymb}
\usepackage{booktabs}
\usepackage{multirow}
\usepackage{makecell}
\usepackage{CJKutf8}
\usepackage{tabularx}
\usepackage{xcolor}
\usepackage[table]{xcolor} 
\usepackage{xspace}
\usepackage{CJKutf8}  
\usepackage{arydshln}
\title{CIRAG: Construction-Integration Retrieval and Adaptive Generation for Multi-hop Question Answering}

\author{
    Zili Wei\thanks{\ \ Equal contribution.}, 
    Xiaocui Yang\footnotemark[1],
    Yilin Wang,
    Zihan Wang, \\
    \textbf{Weidong Bao,
    Shi Feng\thanks{\ \ Corresponding author.},
    Daling Wang,
    Yifei Zhang} \\
    \\ 
    \textsuperscript{1}Northeastern University, China\\
    \small \ttfamily{
    {weizl2@mails.neu.edu.cn}, 
    {yangxiaocui@cse.neu.edu.cn}, 
    \{wangyilin0409,wzh1998921\}@gmail.com}, \\
     \small \ttfamily{
    {2401808@stu.neu.edu.cn}, 
    \{fengshi, wangdaling, zhangyifei1\}@cse.neu.edu.cn
    }
}

\usepackage{tikz}
\usepackage{pgfplots}
\pgfplotsset{compat=1.17}
\usepackage{enumitem}
\usepackage[most]{tcolorbox}
\usepackage{xcolor}
\usepackage{graphicx}  
\usepackage{booktabs}  
\usepackage{multirow}
\begin{document}
\maketitle
\begin{abstract}
Triple-based Iterative Retrieval-Augmented Generation (iRAG)  mitigates document-level noise for multi-hop question answering. However, existing methods still face limitations: (i) \textbf{greedy single-path expansion}, which propagates early errors and fails to capture parallel evidence from different reasoning branches, and (ii) \textbf{granularity-demand mismatch}, where a single evidence representation struggles to balance noise control with contextual sufficiency. In this paper, we propose the Construction-Integration Retrieval and Adaptive Generation model, \textbf{CIRAG}. It introduces an Iterative Construction-Integration module that constructs candidate triples and history-conditionally integrates them to distill core triples and generate the next-hop query. This module mitigates the greedy trap by preserving multiple plausible evidence chains. Besides, we propose an Adaptive Cascaded Multi-Granularity Generation module that progressively expands contextual evidence based on the problem requirements, from triples to supporting sentences and full passages. Moreover, we introduce Trajectory Distillation, which distills the teacher model's integration policy into a lightweight student, enabling efficient and reliable long-horizon reasoning. Extensive experiments demonstrate that CIRAG achieves superior performance compared to existing iRAG methods.
\end{abstract}

\section{Introduction}

Retrieval-Augmented Generation (RAG) excels in simple queries ~\citep{lewis2020retrieval,lin2024ra,ram2023context} but struggles with multi-hop reasoning ~\citep{trivedi2023interleaving,fan2024survey,mallen2023not}, as single-step retrieval often fails to gather interconnected evidence~\citep{shao2023enhancing}. Iterative RAG (iRAG) ~\citep{trivedi2023interleaving, asai2024self, yao2024seakr} is introduced by retrieving information in multiple steps. However, existing iRAG methods, whether retrieving full documents~\cite{zhao2021multi} or generating chain-of-thoughts~\cite{trivedi2023interleaving}, often accumulate irrelevant noise~\cite{yoran2024making} or factual hallucinations during iterations~\cite{wang2023survey,luo2024reasoning}, which ultimately degrades reasoning reliability.

\begin{figure}[tb]
\centering
\includegraphics[width=0.5\textwidth]{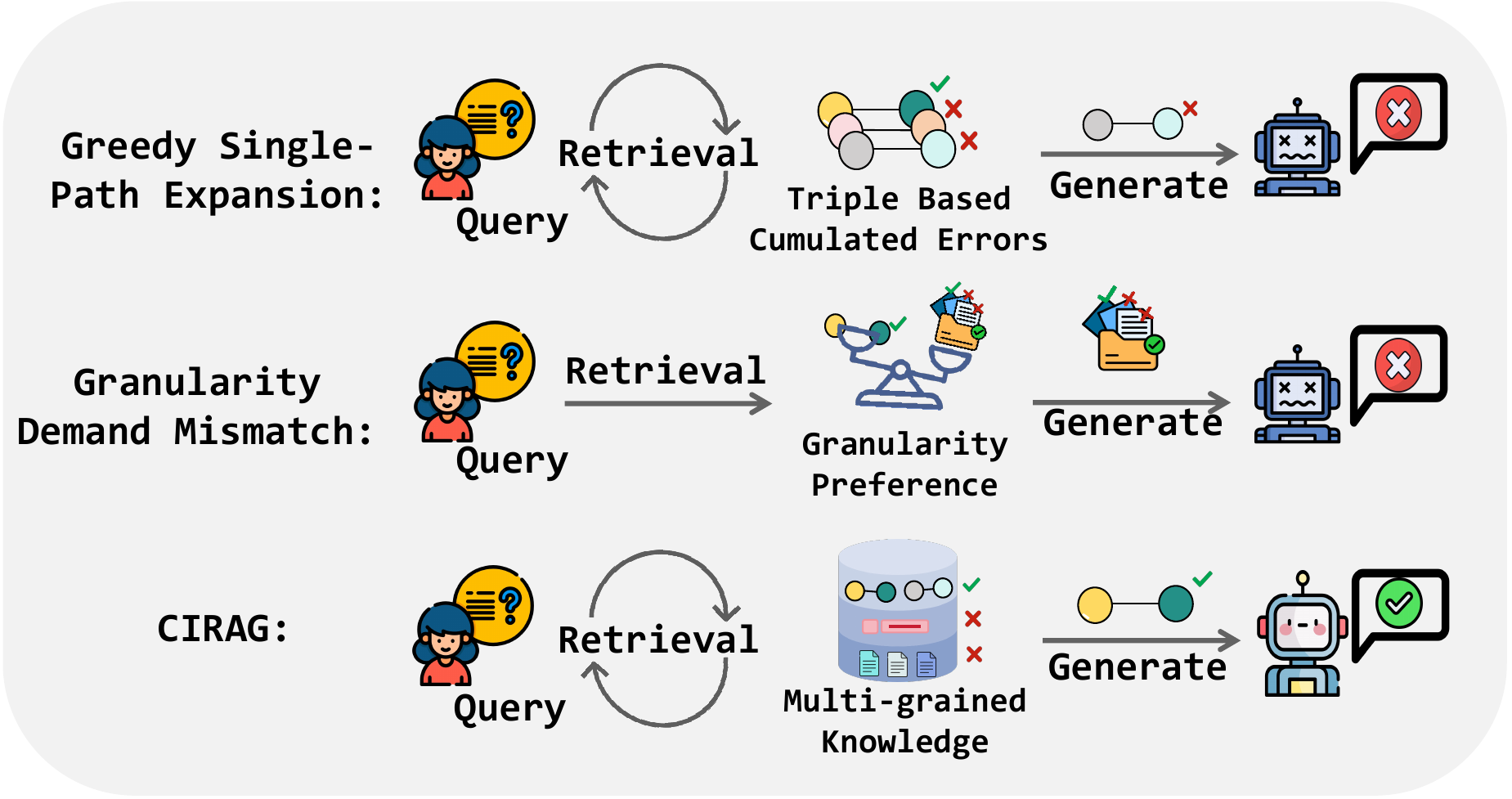}
\vspace{-1.0em}
\caption{Challenges in Triple-based Retrieval.}
\label{figure:intro}
\vspace{-1.em}
\end{figure}

To mitigate this issue, recent research has pivoted towards \textit{triple-based retrieval}~\cite{fang2025kirag,fang2024trace,zhang2025tearag}. By using structured knowledge triples as retrieval units, these methods aim to achieve a more focused and reliable retrieval process. Despite these advances, current triple-based paradigms face two critical limitations, as illustrated in Figure \ref{figure:intro}.
The first challenge is the \textbf{Greedy Single-Path Expansion} in retrieval. Existing methods predominantly select only the single best triple at each step~\cite{fang2024trace,fang2025kirag}. This linear strategy is inherently fragile: minor errors in early decisions can rapidly propagate and compound~\cite{jiapeng2024tree, shi2023large, lee2022generative}. Moreover, by committing to a single path, it overlooks parallel evidence that is often essential for answering complex queries, resulting in incomplete or fragmented reasoning chains~\cite{zhang2024end}.
The second challenge is the \textbf{Granularity-Demand Mismatch}. Current paradigms typically adopt a static evidence representation, failing to account for the heterogeneous information needs of different questions~\cite{fang2024trace,fang2025kirag}. For simple, relation-centric queries, retrieving full documents introduces substantial noise compared to concise triples. Conversely, for complex reasoning tasks that require rich contextual information, structured triples often discard crucial context~\cite{wang2025proprag}, omitting linguistic nuances that are naturally preserved in passages. As a result, a fixed retrieval granularity is insufficient to meet the diverse reasoning requirements posed by different queries.

To address these challenges, we draw inspiration from Construction-Integration (CI) model in cognitive psychology~\cite{kintsch1978toward}. CI  characterizes human comprehension as a two-stage process. In Construction, semantic units are broadly activated, and in the Integration stage, contextual constraints suppress irrelevant activations to yield a coherent semantic network. Grounded in CI, we propose the Construction-Integration Retrieval and Adaptive Generation model, \textbf{CIRAG}, consisting of the Iterative Construction-Integration (ICI) Retrieval module and the Adaptive Cascaded Multi-Granularity Knowledge-Enhanced Generation (ACMG) module.

Specifically, ICI instantiates CI for iterative retrieval to mitigate greedy single-path expansion. At each iteration, the construction phase activates candidate triples extracted from retrieved documents, while the Integration phase enforces global constraints from the accumulated history to suppress noisy or off-path candidates, yielding a core triple set. Based on the uncovered knowledge gap, the integrator further synthesizes the next-hop query to continue the CI loop. By repeatedly alternating between construction and integration, ICI retains the coherent evidence network and reduces single-path bias.
ACMG tackles the granularity-demand mismatch via dynamically expanding context, beginning with compact triples and escalating to supporting sentences or passages only when required, balancing noise control with contextual completeness.
To ensure the core advantages of CIRAG are preserved even in smaller-scale models, we introduce Trajectory Distillation, which transfers integration trajectories from a strong teacher model to a lightweight student, enabling robust multi-step integration with reduced computational overhead. Our contributions can be summarized as follows:


\begin{itemize}[noitemsep, left=0pt]
\item Inspired by the CI model, we propose CIRAG, which effectively addresses the challenges of greedy retrieval bias and mismatched granularity requirements in multi-hop reasoning by synergistically combining ICI and ACMG modules.
\item We propose Trajectory Distillation to transfer integration trajectories from a teacher to an efficient student model, ensuring reasoning capabilities with reduced computational overhead.
\item Extensive experiments on multiple multi-hop and single-hop QA benchmarks validate the effectiveness of CIRAG.
\end{itemize}

\section{Related Works}

\subsection{Text-based iRAG}
The current iRAG methods primarily retrieve relevant text passages from the corpus at each retrieval, providing context for LLM generation. IRCoT ~\cite{trivedi2023interleaving} and Iter-RetGen ~\cite{shao2023enhancing} dynamically generate sub-queries, retrieve relevant documents, and iteratively refine the reasoning trajectory throughout the generation process. FLARE~\cite{jiang2023active} focuses on adaptively retrieving documents when low-probability tokens are generated. MetaRAG ~\cite{zhou2024metacognitive} first generates heuristic answers based on the question and the retrieved documents, and then refines them through retrieval. DualRAG~\cite{cheng2025dualrag} guides retrieval through reason-driven query generation and integrates multi-round retrieval documents with an entity-centric approach. These models perform iterative retrieval by progressively augmenting the query with previously retrieved documents ~\cite{zhao2021multi,trivedi2023interleaving}. However, retrieved documents often include noise or irrelevant information ~\cite{yoran2024making}. The propagation of these distracting contexts can degrade retrieval quality and ultimately hinder overall RAG performance.

\subsection{Triple-based iRAG}
To reduce the impact of noise in documents, the triple-based iterative RAG method improves retrieval granularity by using knowledge triples during the retrieval process~\cite{jimenez2024hipporag,gutierrez2025rag,li2025knowtrace}. KiRAG~\cite{fang2025kirag} decomposes documents into structured triples and gradually expands the knowledge chain composed of triples during iterative retrieval, thereby accurately locating the key information missing in multi-hop question answering. TeaRAG~\cite{zhang2025tearag} employs a triple-enhanced iterative retrieval strategy, simultaneously retrieving text blocks and pre-built triples in each iteration. 
However, existing methods face two challenges: the Greedy Single-Path Expansion and the Granularity-Demand Mismatch. Unlike existing methods, our approach addresses these challenges by obtaining a core set of triples through integration based on historical information in each iteration, and then cascadingly expanding the context to match the most appropriate granularity of information for each question.

\section{CIRAG}
\subsection{Problem Formulation}
We formally define the RAG task: given a user question $x$ and a large-scale document corpus $D = \{d_i\}_{i=1}^N$, the objective of a RAG system is to generate an accurate answer $\hat{a}$ by retrieving and leveraging relevant documents from $D$.

\subsection{Overview}
\label{sec:overview}
As illustrated in Figure \ref{fig:main}, we propose \textbf{CIRAG}, a two-module framework comprising Iterative Construction-Integration (ICI) and Adaptive Cascaded Multi-Granularity Knowledge-Enhanced Generation (ACMG). \textbf{ICI} retrieval module, retrieves the core triples set through two iterative phases: construction phase and integration phase. \textbf{ACMG} module generates the final answer by selecting the most appropriate information for the question through an adaptive cascading method.

\subsection{Iterative Construction-Integration}
\label{sec:Iterative Construction-Integration}

\begin{figure*}[t]
    \centering
    \includegraphics[width=\textwidth]{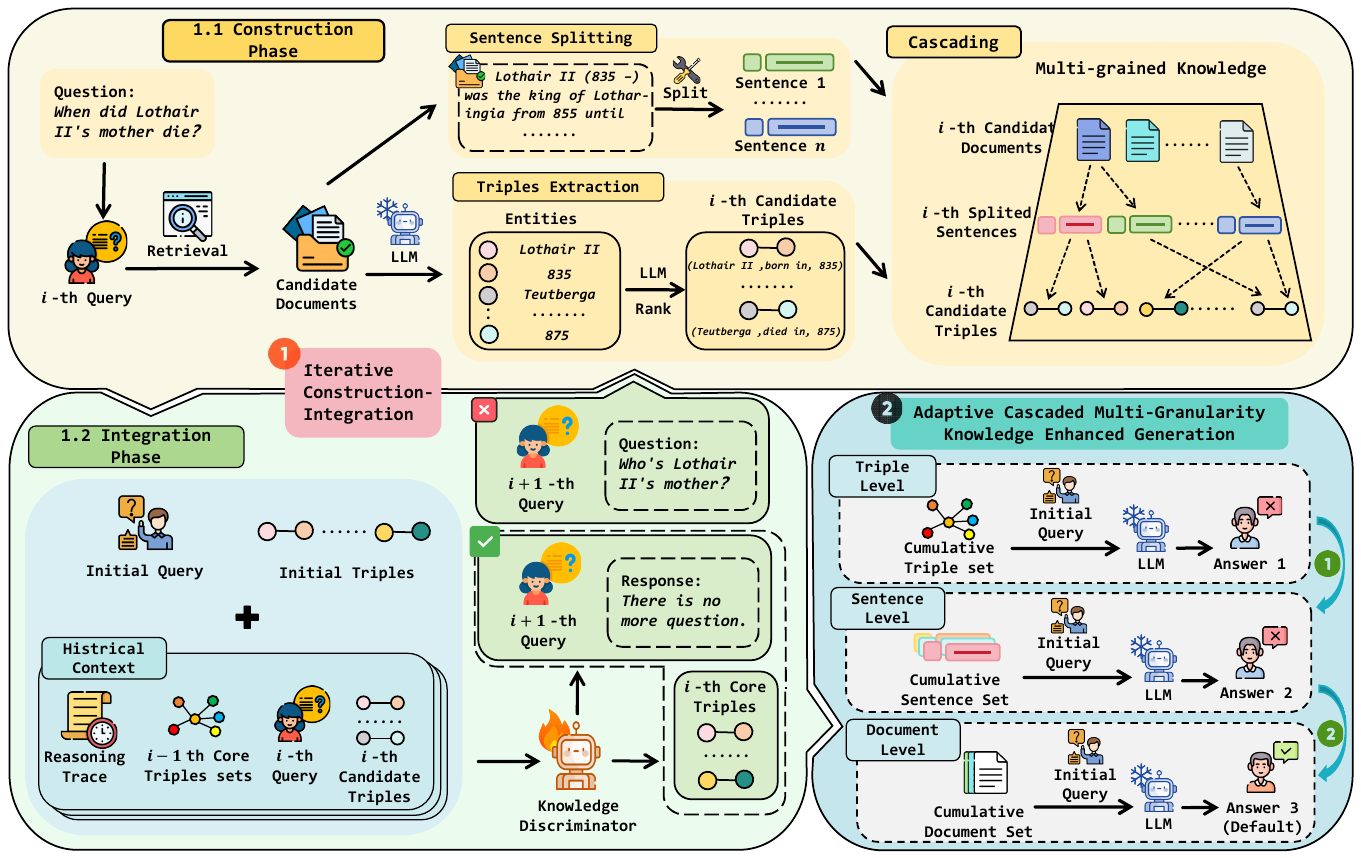}
    \caption{Overview of CIRAG. At each iteration, it employs an Iterative Construction-Integration (ICI) Retrieval module to retrieve a core triple set and record provenance links to their supporting sentences and documents, including two iterative phases: construction phase and integration phase. The core triple set is used to produce the final answer via the Adaptive Cascaded Multi-Granularity Knowledge-Enhanced Generation (ACMG) module.}
    \label{fig:main}
    \vspace{-1em}
\end{figure*}

\paragraph{Construction Phase.}
In the $t$ iteration of ICI (see 1.1 in Figure~\ref{fig:main}), the retriever $\mathcal{R}$ retrieves the top-$K$ documents most relevant to the current query $a_{t}$, forming a document set $\mathcal{D}_t = \{d_t^j\}_{j=1}^K$. These documents constitute the discourse context and are segmented into a unified sentence set $\mathcal{S}_t = \{s_t^k\}_{k=1}^M$. Leveraging a prompt-based approach, for each document in $\mathcal{D}_t$, we employ LLM to identify entities as intermediate anchors, directly guiding the extraction of relational triples\cite{edge2024local,fang2024trace}. The prompt is provided in Appendix~\ref{app:prompt_knowledge_decomposition}. The retriever $\mathcal{R}$ as the reranker first ranks the triples by calculating their semantic similarity to the current query $a_t$. The top-$N$ \footnote{We provide analysis of the effect of $N$ in Appendix~\ref{app:effect_of_n}.} ranked triples are selected to form the candidate triple set $\hat{\mathcal{T}}_t = \{\tau_t^i\}_{i=1}^Q$. To facilitate efficient context mapping, we explicitly record the provenance of each triple by constructing two mapping sets:
$\mathcal{M}_t^{\mathcal{D}} : [Q] \rightarrow [K]$ and $\mathcal{M}_t^{\mathcal{S}} : [Q] \rightarrow [M]$,
where $\mathcal{M}_t^{\mathcal{D}}(i)=j$ and $\mathcal{M}_t^{\mathcal{S}}(i)=k$ indicate that triple
$\tau_t^{i}$ is extracted from document $d_t^{j}$ and sentence $s_t^{k}$.

\paragraph{Integration Phase.}
In the integration phase of iteration $t$ (see 1.2 in Figure~\ref{fig:main}), we employ a discriminative model $\mathcal{K}_D$ to filter candidate triples and generate the next query. In this phase, the model assesses the alignment of candidates with the current query $a_t$, while strictly anchoring to the original question $x$ to maintain global consistency. Simultaneously, it synthesizes the historical context to identify information gaps to formulate a targeted query for the next iteration. To empower $\mathcal{K}_D$ with the capability to execute this complex dynamic reasoning, we optimize it via \textbf{Trajectory Distillation} (detailed in Sec.~\ref{sec:Trajectory Distillation}). Given the original question $x$, the initial candidate set $\hat{\mathcal{T}}_1$, an instruction prompt $I$ provided in Appendix~\ref{app:prompt_Integration_Phase}, and the historical context $\mathcal{H}_{<t}$, $\mathcal{K}_D$ produces (i) a reasoning trace $r_t$, (ii) a filtered core triple set $\tilde{\mathcal{T}}_t$, and (iii) the next-round query $a_{t+1}$:
\begin{equation}
  \label{eq:kd}
  (r_t, \tilde{\mathcal{T}}_t, a_{t+1}) = \mathcal{K}_D(x, \hat{\mathcal{T}}_1, I, \mathcal{H}_{<t}).
\end{equation}

The historical context $\mathcal{H}_{<t}$ encapsulates the reasoning trajectory of previous iterations, including the model output of each round and the corresponding candidate triples. Specifically, for $\mathcal{H}_{<1}=\varnothing$, for $t>1$, it is defined as:
\begin{equation}
  \label{eq:history}
  \mathcal{H}_{<t} = \bigl\{ (r_i, \tilde{\mathcal{T}}_i, a_{i+1}, \hat{\mathcal{T}}_{i+1}) \bigr\}_{i=1}^{t-1}.
\end{equation}

The core triple set $\tilde{\mathcal{T}}_t$ retains only salient information pertinent to the query. To facilitate multi-granularity reasoning, we project these triples back to their source contexts via the provenance mappings, deriving the core sentence set $\tilde{\mathcal{S}}_t$ and document set $\tilde{\mathcal{D}}_t$:
\begin{equation}
  \label{eq:local_mapping}
  \tilde{\mathcal{S}}_t = \{ s_k \in \mathcal{S}_t \mid (i, k) \in \mathcal{M}_t^{\mathcal{S}}, \tau_i \in \tilde{\mathcal{T}}_t \},
\end{equation}
  \begin{equation}
  \label{eq:local_mapping}
  \tilde{\mathcal{D}}_t = \{ d_j \in \mathcal{D}_t \mid (i, j) \in \mathcal{M}_t^{\mathcal{D}}, \tau_i \in \tilde{\mathcal{T}}_t \}.
\end{equation}

Finally, we update the cumulative triple set $\mathbb{C}$, cumulative sentence set $\mathbb{S}$, and cumulative document set $\mathbb{D}$ via the incremental updates $\mathbb{C} \leftarrow \mathbb{C} \cup \tilde{\mathcal{T}}_t$, $\mathbb{S} \leftarrow \mathbb{S} \cup \tilde{\mathcal{S}}_t$, and $\mathbb{D} \leftarrow \mathbb{D} \cup \tilde{\mathcal{D}}_t$. Together, these accumulated sets $\{\mathbb{C}, \mathbb{S}, \mathbb{D}\}$ constitute the multi-granularity context for cascaded generation.

This iterative process terminates when $a_{t+1}=\varnothing$ or the maximum iteration step $L$ is reached. Otherwise, $a_{t+1}$ triggers the retriever to initiate the construction phase at round $t{+}1$. Through this cycle, the system progressively narrows the retrieval space while integrating verified knowledge for multi-hop reasoning.

\subsection{Adaptive Cascaded Multi-Granularity Knowledge Augmented Generation}
\label{sec:Cascaded Multi-Granularity Knowledge Augmented Generation}
To fully leverage the complementary advantages of different granularities of knowledge in multi-hop question answering, we analyze their trade-offs in semantic expressiveness and noise control. Triples have clear structures and minimal noise, but often lack contextual information. Sentences enrich these triples through local context, thereby improving semantic integrity, but inevitably introducing some irrelevant information. Documents provide the most comprehensive global context, but have the most background noise.
Existing approaches typically rely on a single granularity of evidence, making it difficult to balance semantic completeness and noise control. 

To address this, we propose the Adaptive Cascaded Multi-Granularity Augmented Generation (ACMG) module (see 2 in Figure~\ref{fig:main}). We define the hierarchy of context granularities as an ordered sequence $\mathcal{G} = (g_{\mathbb{C}}, g_{\mathbb{S}}, g_{\mathbb{D}})$, where the precedence relation $\prec$, i.e., $g_{\mathbb{C}} \prec g_{\mathbb{S}} \prec g_{\mathbb{D}}$, denotes an increasing order of both semantic coverage and potential noise. This ordering allows the framework to prioritize high-precision, low-noise evidence and escalate to more exhaustive contexts only when the current level is insufficient.

Formally, for each granularity $g \in \mathcal{G}$, the model generates a response $a^{(g)}$ based on the question $x$, the corresponding context $C^{(g)}$, and a sufficiency instruction $I^{(g)}$ provided in Appendix~\ref{app:prompt_Adaptive_Cascaded}. This instruction directs the model $\mathrm{M_R}$ to evaluate the adequacy of $C^{(g)}$ relative to $x$. It produces a valid answer if the information is sufficient, or a predefined refusal response (e.g., Unanswerable) otherwise:
\begin{equation}
    \label{eq:unified_reader}
    a^{(g)} = \mathrm{M_R}\Bigl(x,\; C^{(g)},\; I^{(g)}\Bigr)
\end{equation}
where $C^{(g)}$ is the granularity-specific context selected from the accumulated pools $\{\mathbb{C},\mathbb{S},\mathbb{D}\}$ according to $g\in\mathcal{G}$.

To identify the minimal sufficient granularity, we define a sufficiency indicator
$\mathrm{Suf}(a)\in\{0,1\}$ that checks whether the model answer $a$ is a refusal.
Specifically, $\mathrm{Suf}(a)=0$ if $a$ matches a predefined refusal template,
and $\mathrm{Suf}(a)=1$ otherwise. The system executes a cascaded search strictly following the precedence defined in $\mathcal{G}$, selecting the most concise yet sufficient granularity:
\begin{equation}
    \label{eq:cascade}
    g^{\star} = \min_{\prec} \Bigl\{ g \in \mathcal{G} \mid \mathrm{Suf}\bigl(a^{(g)}\bigr) = 1 \Bigr\}.
\end{equation}
The final output is the answer $a^{(g^{\star})}$ from the first level that satisfies the sufficiency condition. If no level provides a sufficient answer, the system defaults to $a^{(g_{\textsc{doc}})}$ to maximize answerability.

\subsection{Trajectory Distillation}
\label{sec:Trajectory Distillation}

\textsc{CIRAG} is compatible with LLMs of different parameter scales. However, the Integration Phase is non-trivial: at each iteration, the model must (i) filter noisy candidate triples ($\hat{\mathcal{T}}_t \rightarrow \tilde{\mathcal{T}}_t$) and (ii) synthesize a strategic next-hop query ($a_{t+1}$), both conditioned on the accumulated history from previous iterations. Large LLMs are generally more reliable at maintaining such long-horizon consistency, but invoking them in an interactive loop incurs substantial computational overhead. In contrast, lightweight models are more efficient but often fail to produce stable filtering and query-planning decisions.
To achieve an efficient yet reliable integrator, we propose \textbf{Trajectory Distillation}. The core idea is to distill interactive trajectories produced by a strong teacher model into a lightweight student model $\mathcal{K}_D$(Eq.~\eqref{eq:kd}), so that the student can reproduce the teacher's stepwise integration decisions at a lower cost.

\paragraph{Teacher trajectory generation.}
A trajectory $\xi$ records a sequence of interaction steps. Given a question $x$, an initial candidate triple set $\hat{\mathcal{T}}_{1}$, and an instruction prompt $I$, the teacher generates:
\begin{equation}
\label{eq:traj}
\xi = \bigl\{(y_t, o_{t+1})\bigr\}_{t=1}^{L_\xi} \sim \pi_T(\cdot \mid x, \hat{\mathcal{T}}_{1}, I),
\end{equation}
where at step $t$ the teacher produces an integration decision:
\[
y_t = (r_t, \tilde{\mathcal{T}}_t, a_{t+1}),
\]
consisting of an integration rationale $r_t$, the filtered core triple set $\tilde{\mathcal{T}}_t$, and the next-hop query $a_{t+1}$. The retriever then returns the subsequent observation:
\[
o_{t+1} = \hat{\mathcal{T}}_{t+1} = \mathcal{R}(a_{t+1}),
\]
which serves as a candidate set for the next iteration.

\paragraph{Student supervision.}
Following prior works \cite{chen2023fireact,gou2023tora,kang2025distilling}, we fine-tune the student to predict the teacher's integration decisions, while treating retrieval observations as context rather than supervision targets:
\begin{equation}
\label{eq:distill_obj}
\min_{\theta}\;-\mathbb{E}_{\substack{x\sim \mathcal{D}_{\text{train}} \\ \tau\sim \pi_T(\cdot\mid x,I)}}
 \sum_{t=1}^{L}\log \mathcal{K}_D \bigl( y_t \mid x, I, \tau_{<t}; \theta \bigr)
\end{equation}
where $\tau_{<t}=\{(y_i, o_i)\}_{i=1}^{t-1}$. \(\mathcal{D}_{\text{train}}\) denotes the training set, \(\pi_T(\cdot\mid x,I)\) is the teacher trajectory distribution conditioned on input \(x\) and prompt \(I\), \(\mathcal{K}_D(\cdot;\theta)\) is the student model parameterized by \(\theta\), and \(L_{\tau}\) is the length of trajectory \(\tau\). At each step \(t\), the student predicts the reasoning trace \(r_t\), the filtered triples \(t_t\), and the next query \(a_t\), conditioned on \(x\) and the trajectory history \(\tau_{<t}\).

 After distillation, $\mathcal{K}_D$ can execute the same integration loop by consistently selecting salient triples ($\tilde{\mathcal{T}}_t$) and synthesizing next-hop queries ($a_{t+1}$) across iterations, while remaining substantially more efficient than the teacher.
\begin{table*}[t]
    \centering
    \resizebox{0.9\linewidth}{!}{%
\begin{tabular}{lcccccccccccc}
\toprule
\multirow{3}{*}{\textbf{Method}} & \multicolumn{6}{c}{\textbf{Qwen2.5-7B}} & \multicolumn{6}{c}{\textbf{Qwen2.5-max}} \\
\cmidrule(lr){2-7} \cmidrule(lr){8-13}
 & \multicolumn{2}{c}{\textbf{2WikiMQA}} & \multicolumn{2}{c}{\textbf{HotpotQA}} & \multicolumn{2}{c}{\textbf{MuSiQue}} & \multicolumn{2}{c}{\textbf{2WikiMQA}} & \multicolumn{2}{c}{\textbf{HotpotQA}} & \multicolumn{2}{c}{\textbf{MuSiQue}} \\
\cmidrule(lr){2-3} \cmidrule(lr){4-5} \cmidrule(lr){6-7} \cmidrule(lr){8-9} \cmidrule(lr){10-11} \cmidrule(lr){12-13}
 & \textbf{F1} & \textbf{EM} & \textbf{F1} & \textbf{EM} & \textbf{F1} & \textbf{EM} & \textbf{F1} & \textbf{EM} & \textbf{F1} & \textbf{EM} & \textbf{F1} & \textbf{EM} \\
\midrule
NativeRAG & 31.5 & 28.2 & 50.2 & 34.5 & 16.8 & 9.9 & 47.1 & 39.2 & 66.9 & 52.4 & 27.6 & 18.2 \\
IRCOT & 45.7 & 36.4 & 56.8 & 42.4 & 23.5 & 13.4 & 65.7 & 55.3 & 72.8 & 58.1 & 34.2 & 22.1 \\
FLARE & 43.1 & 35.2 & 56.4 & 41.9 & 23.8 & 13.5 & - & - & - & - & - & - \\
MetaRAG & 50.4 & 44.7 & \underline{63.3} & \underline{49.6} & 31.9 & 21.2 & 58.7 & 52.4 & \underline{74.6} & \underline{60.8} & 43.8 & 32.4 \\
KiRAG & 52.7 & 36.9 & 62.1 & 49.0 & 31.7 & 20.2 & 59.4 & 52.1 & 73.2 & 59.0 & 45.0 & 32.7 \\
DualRAG & 62.3 & 51.7 & 58.7 & 44.8 & 33.7 & 22.1 & \underline{75.6} & \underline{65.8} & 73.3 & 57.8 & \underline{50.2} & \underline{36.6} \\
DualRAG-FT & \underline{65.6} & \underline{53.8} & 62.6 & 47.1 & \underline{35.8} & \underline{25.1} & - & - & - & - & - & - \\
\midrule
\textbf{Ours} & \textbf{69.5} & \textbf{59.0} & \textbf{67.1} & \textbf{52.5} & \textbf{40.9} & \textbf{29.3} & \textbf{76.4} & \textbf{67.1} & \textbf{74.9} & \textbf{60.9} & \textbf{56.0} & \textbf{44.9} \\
\bottomrule
\end{tabular}%
}
  \caption{Results on three MHQA benchmarks with Qwen2.5-7B-Instruct and Qwen2.5-max as base LLMs. \textbf{Bold} marks the best and \underline{underline} the second-best.}
  \label{table:fffn}
\end{table*}

\begin{table}[t]
\centering
\small
\setlength{\tabcolsep}{3pt}
\renewcommand{\arraystretch}{1.1}
\begin{tabular*}{\columnwidth}{@{\extracolsep{\fill}}lcccccc}
\toprule
\multirow{2}{*}{\textbf{Method}} &
\multicolumn{2}{c}{\textbf{2WikiMQA}} &
\multicolumn{2}{c}{\textbf{HotpotQA}} &
\multicolumn{2}{c}{\textbf{MuSiQue}} \\
\cmidrule(lr){2-3} \cmidrule(lr){4-5} \cmidrule(lr){6-7}
& \textbf{F1} & \textbf{EM} & \textbf{F1} & \textbf{EM} & \textbf{F1} & \textbf{EM} \\
\midrule
\textbf{CIRAG} & \textbf{68.1} & \textbf{58.9} & \textbf{67.1} & \textbf{52.5} & \textbf{40.9} & \textbf{29.3} \\
w/o TD       & 59.7 & 49.8 & 62.5 & 48.8 & 33.1 & 22.2 \\
w/o reranker       & 66.3 & 57.3 & 64.7 & 50.6 & 38.3 & 27.2 \\
\bottomrule
\end{tabular*}

\caption{Ablation Study on trajectory distillation and reranker for CIRAG
Using Qwen2.5-7B-Instruct.}
\label{tab:ablation_td}

\end{table}

\section{Experiment}
\subsection{Datasets and Metrics}
We evaluate our method on three multi-hop QA benchmarks: \textbf{HotpotQA}~\cite{yang2018hotpotqa}, \textbf{2WikiMultiHopQA}(\textbf{2WikiMQA})~\cite{ho2020constructing}, and \textbf{MuSiQue}~\cite{trivedi2022musique}. For each dataset, we randomly selected 1000 multi-hop questions for evaluation in the validation set as done in previous work~\cite{trivedi2023interleaving}. To create a more rigorous and realistic retrieval setting, we followed the IRCoT~\cite{trivedi2023interleaving} settings and merged all supported and unsupported paragraphs from the selected questions in each dataset to build the retrieval database.
We use Exact Match (EM) and F1 as evaluation metrics, which are the standard metrics for these datasets. More details can be found in Appendix~\ref{app:experimental_details}.

\subsection{Baselines}We propose a simple and efficient Iterative Retrieval-Augmented Generation framework, which we mainly compare with the iterative RAG method. Specifically, it mainly includes the following categories of RAG methods: (i) NativeRAG ~\cite{lewis2020retrieval}, which follows a retrieval-generation paradigm and generates answers based on documents retrieved once. (ii) Text-based iterative RAG methods, which iteratively retrieve relevant documents to gather the key information needed for multi-hop reasoning, such as IRCoT~\cite{trivedi2023interleaving}, FLARE~\cite{jiang2023active}, MetaRAG~\cite{zhou2024metacognitive}, DualRAG and its variant DualRAG-FT~\cite{cheng2025dualrag}. (iii) Triple-based iterative RAG method, which efficiently supplies the knowledge needed for multi-hop reasoning, using triples as retrieval units, such as KiRAG~\cite{fang2025kirag}. More details can be found in Appendix~\ref{app:Baselines}

\subsection{Implementation and Training Details}
\paragraph{Backbone.} We use Qwen-2.5-7B-Instruct and Qwen-max-2025-01-25~\cite{yang2024qwen2} as the backbone of our framework and all baselines.

\paragraph{Retrieval Setup.} We use two different retrieval models to validate the compatibility of our approach, including bge-Small-env1.5~\cite{xiao2024c} and nvidia/NVEmbed-v2~\cite{lee2024nv}. In the main experiment, all iterative RAG methods are set to a maximum of 4 iteration steps, and the retriever was used to retrieve the top 10 documents for each question for model inference. We provide the analysis of the effect of iteration steps in Appendix~\ref{app:effect_of_L}.

\paragraph{Trajectory Distillation.} Using Qwen-max-2025-01-25 as the teacher model, we apply CIRAG to generate 3,000 complete reasoning trajectories
from the training sets of HotpotQA~\cite{yang2018hotpotqa}, 2WikiMultihopQA~\cite{ho2020constructing}, and MuSiQue~\cite{trivedi2022musique}. We fine-tune Qwen-2.5-7B-Instruct as the student model utilizing Low-Rank Adaptation (LoRA)~\cite{hu2022lora}.
Further training configurations and implementation details are provided in Appendix~\ref{app:training_hparam_details}.

\begin{figure}[tb]
\centering
\includegraphics[width=0.5\textwidth]{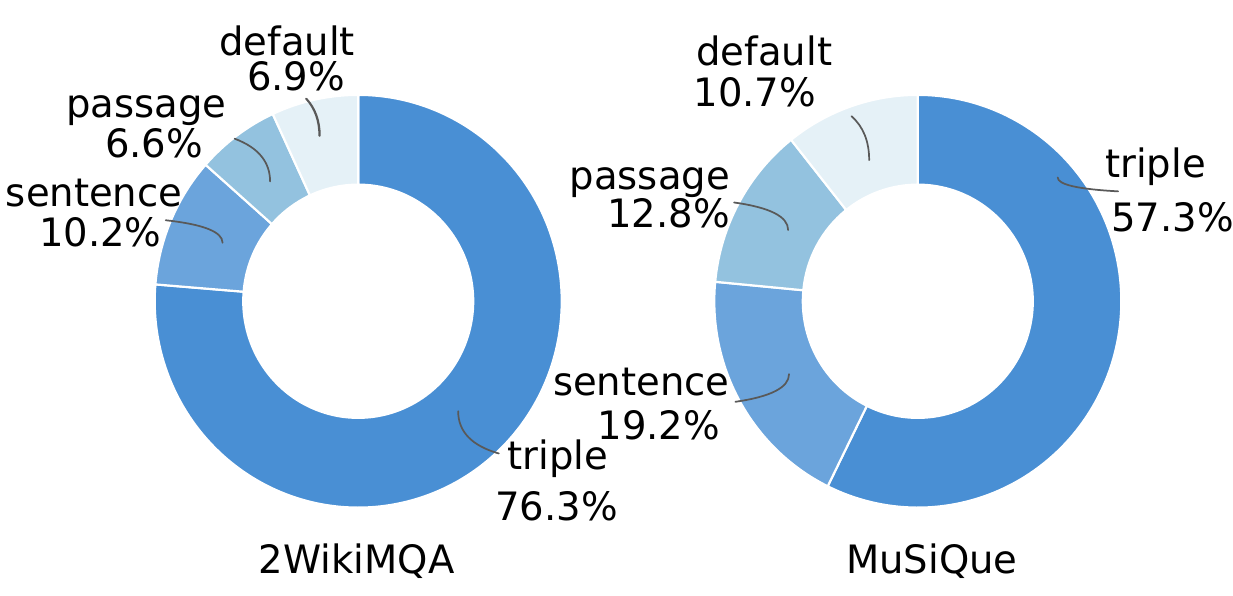}
\vspace{-1.0em}
\caption{Context granularity distribution across datasets}
\label{figure:granularity_distribution}
\vspace{-1.em}
\end{figure}

\subsection{Main Results}
Table~\ref{table:fffn} reports the main experimental results on three standard MHQA benchmarks. Overall, \textbf{CIRAG} consistently outperforms all baselines across varying model scales. We make the following key observations:
(1) Compared with text-based iRAG methods, \textbf{CIRAG} improves performance by an average of 4.3\% (F1) and 4.9\% (EM) on Qwen2.5-7B-Instruct. Notably, KiRAG, despite optimizing only the triple-retrieval component, remains competitive with stronger text-based approaches, e.g., MetaRAG and DualRAG. These results suggest that triples serve as finer-grained retrieval units than full passages, enabling more accurate, stable iterative retrieval, which ultimately benefits QA performance.
(2) Compared with triple-based iRAG methods, \textbf{CIRAG} achieves an average gains of 10.3\% (F1) and 11.6\%(EM) with Qwen2.5-7B-Instruct. These improvements validate our design, yielding a better balance between structured precision and semantic completeness. The history-conditioned integration step mitigates the single-path bias of greedy linear expansion, while matching the context granularity to the question requirements. 
(3) Figure~\ref{figure:granularity_distribution} reports the Granularity Distribution on 2WikiMQA and MuSiQue. We observe that triples dominate the cascade, accounting for 76.3\% on 2WikiMQA and 57.3\% on MuSiQue. This indicates that most multi-hop questions can be resolved with compact relational triples, allowing to avoid unnecessary noise. Meanwhile, the distribution shifts with dataset difficulty: in the more challenging MuSiQue dataset, the usage of coarser granularities is higher compared to 2WikiMQA. This contrast highlights the heterogeneous granularity demanded by different questions. Overall, these two observations confirm the rationale of our method, which can remain at low-noise triples when sufficient, yet reliably escalate to sentences or passages when additional context is required.
(4) To verify the effectiveness of CIRAG, we report additional results under different retrievers and backbones in Appendix~\ref{app:overall_performance_different_retrievers_readers}.

\begin{figure}[tb]
\centering
\includegraphics[width=0.45\textwidth]{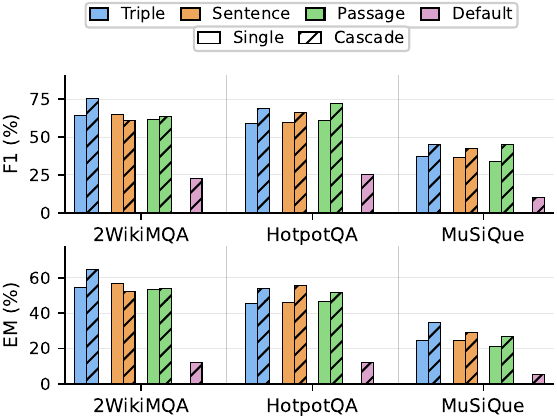}
\vspace{-0.5em}
\caption{Comparison of single-granularity and cascaded performance}
\vspace{-0.5em}
\label{figure:cascade_vs_single_with_default_no_numbers_no_overlap}
\end{figure}

\subsection{Ablation Study}
\paragraph{Effect of Evidence Granularity and Cascaded Context Expansion} Table~\ref{tab:main_results} compares different evidence granularities and cascade variants on Qwen2.5-7B-Instruct. We observe strong task dependency for single-granularity settings. On 2WikiMQA and MuSiQue, w/o Triple + Sentence underperform the others, suggesting that longer contexts introduce distracting noise; in contrast, passages are more effective on HotpotQA, indicating a higher need for paragraph-level context to recover bridging evidence. This contrast highlights heterogeneous granularity demands in multi-hop QA. Across all datasets, two-level cascades consistently outperform single-granularity baselines, showing that cascading can expand context when necessary while avoiding redundant context injection. \textsc{CIRAG} achieves the best overall F1/EM, confirming the benefit of full cascaded expansion.
We further compare performance at each cascade stage with its single-granularity counterpart in Figure~\ref{figure:cascade_vs_single_with_default_no_numbers_no_overlap} . The cascade improves F1/EM consistently at every granularity, indicating that on-demand expansion selects more suitable evidence for different questions and yields more reliable QA performance.

\begin{figure}[tb]
\centering
\includegraphics[width=0.39\textwidth]{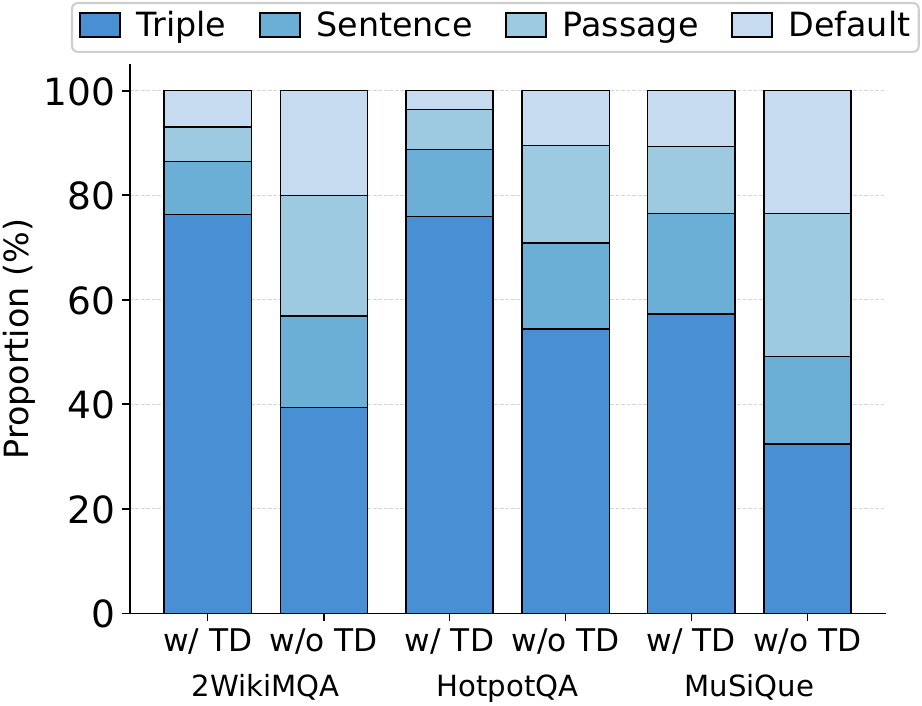}
\vspace{-0.5em}
\caption{Effect of Trajectory Distillation on context granularity distribution }
\label{figure:Granularity_distribution_DT}
\end{figure}

\begin{table}[t] 
\centering
\resizebox{\columnwidth}{!}{%
\begin{tabular}{lcccccc}
\toprule
\multirow{2}{*}{\textbf{Method}} & \multicolumn{2}{c}{\textbf{2WikiMQA}} & \multicolumn{2}{c}{\textbf{HotpotQA}} & \multicolumn{2}{c}{\textbf{MuSiQue}} \\
\cmidrule(lr){2-3} \cmidrule(lr){4-5} \cmidrule(lr){6-7}
 & \textbf{F1} & \textbf{EM} & \textbf{F1} & \textbf{EM} & \textbf{F1} & \textbf{EM} \\
\midrule
\textbf{CIRAG (Full)} & \textbf{68.1} & \textbf{58.9} & \textbf{67.0} & \textbf{52.6} & \textbf{40.9} & \textbf{29.5} \\
\midrule
w/o Passage & 67.4 & 57.5 & 65.5 & 51.1 & 40.8 & 28.9 \\
w/o Triple & 67.5 & 58.8 & 64.9 & 51.0 & 38.7 & 26.9 \\
\midrule
w/o Sentence + Passage & 64.1 & 54.7 & 59.1 & 45.4 & 37.1 & 24.5 \\
w/o Triple + Passage & 64.7 & 56.5 & 59.7 & 46.1 & 36.7 & 24.2 \\
w/o Triple + Sentence & 61.9 & 53.2 & 61.1 & 46.5 & 34.1 & 21.1 \\
\bottomrule
\end{tabular}%
}
\small 
\caption{Ablation Study on Cascaded Multi-Granularity Knowledge Augmented strategies for CIRAG
Using Qwen2.5-7B-Instruct.}
\vspace{-0.5em}
\label{tab:main_results}
\vspace{-0.5em}
\end{table}

\paragraph{Effect of Trajectory Distillation}
To assess the impact of Trajectory Distillation (TD) on the \textit{Integration Phase}, we implement a variant, w/o TD, where the distilled model $\mathcal{K}_D$ is replaced by a frozen Qwen-2.5-7B base model. As shown in Table~\ref{tab:ablation_td}, removing TD results in a significant degradation in QA performance. 
The decline is further elucidated by the granularity distribution in Figure~\ref{figure:Granularity_distribution_DT}. The utilization ratio of triples in the cascaded process markedly decreases, while the incidence of passages and default (where no matching granularity is found) surges. This implies a substantial deterioration in the quality of the core triple set retrieved by the ICI loop, which directly impairs downstream generation. These findings substantiate the critical role of TD in endowing the model with the capability to discriminate core triples and orchestrate multi-hop queries.

\paragraph{Effect of Reranker}
To assess the impact of the reranker, we introduce a variant, w/o reranker, which treats all triples extracted from retrieved documents as candidate triples. As shown in Table~\ref{tab:ablation_td}, this variant achieves comparable performance across all datasets, validating the effectiveness of the \textit{Integration Phase} in identifying relevant triples.



\begin{figure}[tb]
\centering
\includegraphics[width=0.41\textwidth]{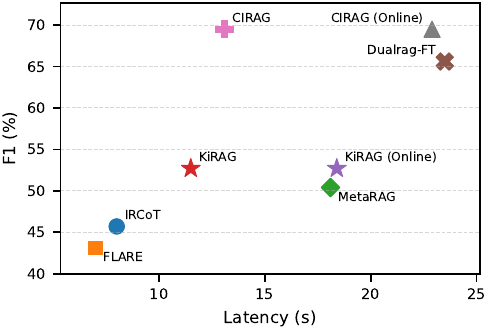}
\vspace{-0.5em}
\caption{Latency vs. F1 on 2WikiMQA Using Qwen2.5-7B-Instruct.}
\label{figure:efficiency_analysis}
\vspace{-0.5em}
\end{figure}

\subsection{Efficiency Analysis}

We evaluate the efficiency of CIRAG compared to baseline methods. To ensure a fair comparison, both CIRAG and the baselines utilize the identical retriever for document retrieval and employ the same underlying model as the reasoning component. Drawing inspiration from prior work~\cite{fang2025kirag}, CIRAG extracts and caches knowledge triples offline to minimize online computational overhead. To quantify the impact of these pre-computed triples, we introduce a variant named CIRAG (Online), which performs dynamic triple extraction during the iterative retrieval process. Hardware configurations. 

Figure~\ref{figure:efficiency_analysis} illustrates the average inference latency versus F1 score on the 2WikiMQA test set. The results indicate that: (1) Compared to CIRAG (Online), the offline pre-computation significantly reduces inference latency with negligible performance degradation, validating the efficiency benefits of our caching strategy. (2) CIRAG achieves higher F1 scores while maintaining relatively low latency, demonstrating that our approach enhances QA accuracy without introducing substantial computational costs. (3) Overall, CIRAG strikes a superior balance between effectiveness and efficiency, characterized by lower latency and higher F1 scores compared to baselines.

\begin{table}[t]
\centering
\footnotesize
\setlength{\tabcolsep}{6pt} 

\begin{tabular}{lcccc}
\toprule
\multirow{2}{*}{\textbf{Method}} & \multicolumn{2}{c}{\textbf{NQ}} & \multicolumn{2}{c}{\textbf{WebQ}} \\
\cmidrule(lr){2-3} \cmidrule(lr){4-5}
 & \textbf{F1} & \textbf{EM} & \textbf{F1} & \textbf{EM} \\
\midrule
Native     & 59.6 & 43.7 & 42.1 & 27.8 \\
IRCOT      & 56.7 & 41.0 & 41.6 & 25.0 \\
FLARE      & 56.3 & 43.0 & 42.3 & 26.0 \\
MetaRAG    & 61.1 & 47.8 & 48.2 & 33.8 \\
Kirag      & 57.1 & 41.7 & 44.6 & 27.5 \\
Dualrag    & 58.2 & 44.2 & 45.1 & 29.5 \\
Dualrag-FT & 60.2 & 45.7 & 47.4 & 31.2 \\
\midrule
\textbf{CIRAG} & \textbf{61.4} & \textbf{46.0} & \textbf{48.3} & \textbf{34.0} \\
\bottomrule
\end{tabular}
\caption{Additional results on single-hop QA datasets using Qwen2.5-7B-Instruct.}
\label{tab:additional}
\vspace{-0.5em}
\end{table}

\subsection{Other QA Tasks}

To evaluate generalization, we conducted additional experiments on a multi-hop QA dataset, WebQA~\cite{berant2013semantic}, and a single-hop QA dataset, NQ~\cite{kwiatkowski2019natural}. As shown in Table~\ref{tab:additional}, CIRAG outperforms all baselines on WebQA and NQ, demonstrating strong generalization across QA settings.

\subsection{Case Study}
We conduct a case study in Appendix~\ref{appendix:case} to verify the effectiveness of our method.

\section{Conclusion}
We propose \textsc{CIRAG}, a two-module iRAG framework for multi-hop QA that integrates triple-based retrieval with adaptive multi-granularity generation. An Iterative Construction-Integration module distills core triples and plans next-hop queries to mitigate greedy reasoning and retrieval noise, while an Adaptive Cascaded Generation module dynamically expands context from triples to sentences and passages as needed. We further introduce Trajectory Distillation to enhance the integration capability of lightweight models. Experiments on MHQA benchmarks demonstrate consistent improvements over iRAG baselines. In the future, we will explore lightweight routing for efficient granularity selection.

\section*{Limitations}
Our framework has two main limitations. First, the Construction Phase depends on prompt-based open IE; in specialized domains or when relations are highly implicit, the extractor may miss key triples, limiting downstream integration. Future work could improve extraction via domain-adaptive training or prompt optimization. Second, the cascaded generation incurs extra latency due to sequential granularity checks. While Trajectory Distillation reduces integration cost, generation overhead remains. A potential improvement is to develop a lightweight granularity router that directly selects the appropriate context level, which could further accelerate inference.

\bibliography{anthology,custom}

\appendix
\section{Prompts}

\subsection{Prompt for Knowledge Triple Extraction}
\label{app:prompt_knowledge_decomposition}

The prompt used for NER from a document is illustrated in Figure~\ref{figure:prompt_NER}. The prompt used for extracting knowledge triples from a document is illustrated in Figure~\ref{figure:prompt_knowledge_triple_extraction}. 

\subsection{Prompt for Integration Phase}
\label{app:prompt_Integration_Phase}

The prompt used for distilling a core triple set and synthesizing the next-hop query in the integration phase is illustrated in Figure~\ref{figure:prompt_Integration_Phase}. 

\subsection{Prompt for ACMG}
\label{app:prompt_Adaptive_Cascaded}

Figure~\ref{figure:prompt_ACMG} illustrates the prompting method at the ACMG triplet level. The sentence and article levels differ from the triplet level only in the examples used.

\section{Experimental Details}
\label{app:experimental_details}

\subsection{Datasets}
\label{app:datasets}
In our experiments, we employ four multi-hop QA datasets: HotpotQA, 2WikiMultiHopQA, MuSiQue, WebQuestions (WebQA), and one single-hop QA dataset: Natural Questions (NQ).
For HotpotQA, 2WikiMultiHopQA, and MuSiQue, we construct the retrieval corpus by following exactly the same procedure as \citet{trivedi2023interleaving}.
For WebQA and NQ, we use the corpus version released with DPR.
To control evaluation cost, for each question in WebQA and NQ, we include all annotated supporting documents and additionally sample up to 10 non-supporting documents.

For datasets with public test sets (WebQA and NQ), we randomly sample 500 test questions for evaluation.
For datasets without public test sets (HotpotQA, 2WikiMultiHopQA, and MuSiQue), we randomly sample 1{,}000 questions from the development set as our test split and report performance on this subset.
Since these three datasets are also used for training, we further randomly sample 1{,}000 questions from each original training set to form the training split used in our experiments.

\subsection{Metrics Details}
\label{app:Metrics}
Exact Match (EM) provides the strictest criterion, assigning a score of 1 only when the predicted answer exactly matches the ground truth and 0 otherwise. 

The F1 score measures token-level similarity by computing the harmonic mean of Precision and Recall, where Precision reflects the proportion of predicted tokens that are correct, and Recall denotes the proportion of reference tokens successfully retrieved. We follow evaluation metrics from MuSiQue~\cite{trivedi2022musique} to calculate F1 scores for the final answer.

\subsection{Baselines}
\label{app:Baselines}
For IRCoT and FLARE, we use the implementations provided by DualRAG~\cite{cheng2025dualrag}. For other models, we adapt the official implementations released by the authors to match our experimental setting.
For a fair comparison, CIRAG and all baselines use the same retriever to access the same corpus and share the same backbone LLM for reasoning and answer generation.
For methods that require training but do not provide public checkpoints, we reproduce training by following the authors' released data-generation pipelines to construct the training set, and we use the same initial training split as CIRAG.
When applicable, we align the training recipe (e.g., optimization settings and hyperparameters) with the original papers as closely as possible under our hardware constraints.

\subsection{Training and Hyperparameter Details}
\label{app:training_hparam_details}

\vspace{0.5em} \noindent \textbf{Training Details.}
We fine-tune student models using parameter-efficient tuning with LoRA (rank 64) ~\cite{hu2022lora}. models are fine-tuned for 2 epochs using a batch size of 2 and a learning rate of 2 * $10^{-4}$. Experiments are conducted using four NVIDIA A6000 48GB GPUs,with a total training time of approximately 3 hours.

\vspace{0.5em} \noindent \textbf{Implementation and Hyperparameter Details.}
Throughout the experiments, we set the maximum number of iterative steps L to 4. The details of each component in our CIRAG are outlined as follows:
For the \textit{Retriever} model, we use either bge-Small-env1.5~\cite{xiao2024c} or nvidia/NVEmbed-v2~\cite{lee2024nv} to retrieve documents and rerank triples. The number of retrieved documents per iteration (i.e., $K$) is $10$. The number of candidate triples per iteration (i.e., $N$) is $30$. 
For the \textit{Backbone} model, we try different LLMs, including Llama3~\cite{dubey2024llama}, Qwen-2.5-7B-Instruct, and Qwen-max-2025-01-25~\cite{yang2024qwen2} as the backbone of our framework and all baselines. We set the temperature to 0 when calling the API of Qwen-max and use greedy decoding for other models to avoid random sampling~\cite{renze2024effect}. We mainly report the performance of using Qwen-2.5-7B-Instruct and Qwen-max-2025-01-25~\cite{yang2024qwen2} as the Backbone.
Hardware configurations: All experiments are conducted on the same hardware environment equipped with an Intel Xeon Gold 6326 CPU (2.90GHz, 32 cores) and an NVIDIA RTX A6000 GPU.

\section{Additional Experimental Results and Analysis}
\label{app:additional_results_and_analysis}

\subsection{Using Different Retrievers and Readers}
\label{app:overall_performance_different_retrievers_readers}
To validate the effectiveness of CIRAG, we provide additional results using different retrievers and a backbone model. 
Specifically, we replace the nvidia/NVEmbed-v2 Retriever with bge-Small-env1.5 Retriever for retrieving documents from the corpus and the other components remain unchanged. The corresponding QA performance is presented in Table~\ref{table:retrieval_performance_bge}, respectively. 
The results are consistent with those obtained using thenvidia/NVEmbed-v2 Retriever, demonstrating the adaptability and effectiveness of our CIRAG across different retriever models. 

To verify the applicability of our approach across different LLM architectures, we also conducted experiments on other open-source models (Llama-3-8B-Instruct). The results are shown in Table~\ref{table:qa_performance_e5_different_readers}. These experimental results indicate that our method is also applicable to other LLM architectures, demonstrating robust performance across different models. 

\begin{figure}[tb]
\centering
\includegraphics[width=0.5\textwidth]{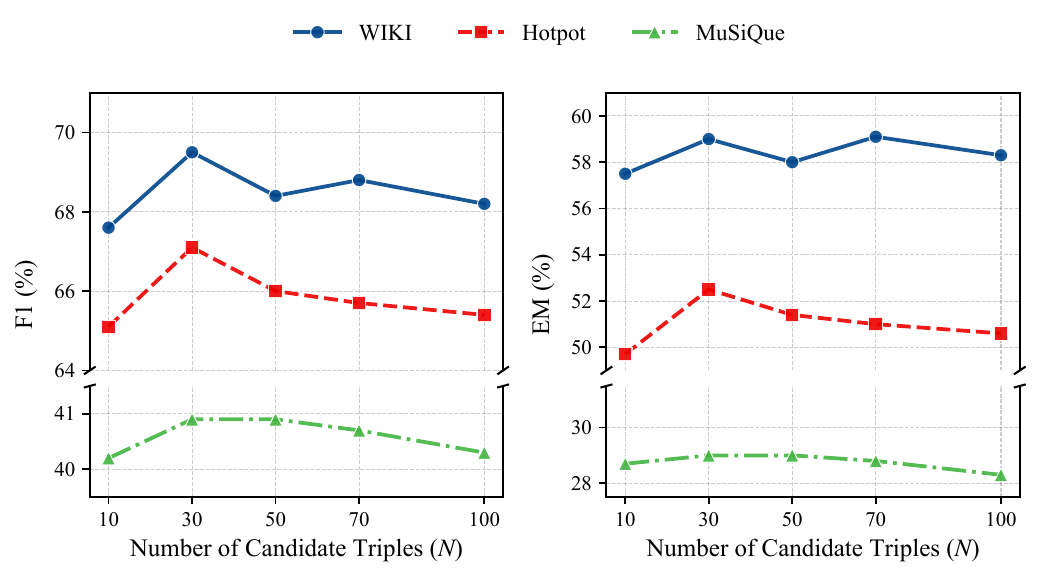}
\vspace{-0.5em}
\caption{QA performance (\%) of CIRAG under different values of $N$ on three multi-hop QA datasets.}
\label{figure:effect_num_candidate_triples}
\vspace{-0.5em}
\end{figure}

\begin{figure*}[tb]
\centering
\includegraphics[width=\linewidth,height=.3\textwidth]{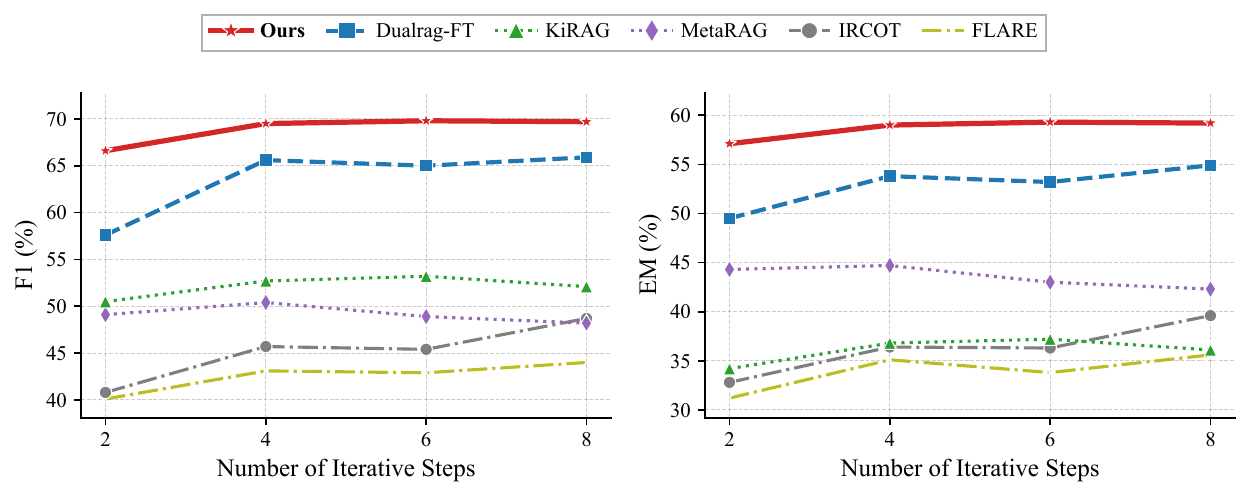}
\vspace{-0.5em}
\caption{The effect of the number of iterative steps $L$ for different models on the 2WikiMQA test set.}
\label{figure:effect_L}
\vspace{-0.5em}
\end{figure*}

\begin{table}[t] 
\centering
\resizebox{\columnwidth}{!}{%
\begin{tabular}{lcccccc}
\toprule
\multirow{2}{*}{\textbf{Method}} 
& \multicolumn{2}{c}{\textbf{2WikiMQA}} 
& \multicolumn{2}{c}{\textbf{HotpotQA}} 
& \multicolumn{2}{c}{\textbf{MuSiQue}} \\
\cmidrule(lr){2-3} \cmidrule(lr){4-5} \cmidrule(lr){6-7}
& \textbf{F1} & \textbf{EM} 
& \textbf{F1} & \textbf{EM} 
& \textbf{F1} & \textbf{EM} \\
\midrule
IRCOT       & 41.1 & 23.1 & 56.7 & 41.3 & 23.1 & 13.6 \\
FLARE       & 40.5 & 24.8 & 56.8 & 41.2 & 24.5 & 14.0 \\
MetaRAG     & 44.9 & 30.6 & \underline{62.5} & \underline{48.5} & 31.7 & 20.8 \\
Kirag       & 47.2 & 20.2 & 60.5 & 46.2 & 30.9 & 19.6 \\
Dualrag     & 56.8 & 37.6 & 57.5 & 43.6 & 32.4 & 21.1 \\
Dualrag-FT  & \underline{60.1} & \underline{39.7} & 61.9 & 46.0 & \underline{34.6} & \underline{24.7} \\
\midrule
\textbf{Ours} 
            & \textbf{64.0} & \textbf{44.9} 
            & \textbf{66.7} & \textbf{51.4} 
            & \textbf{40.7} & \textbf{28.2} \\
\bottomrule
\end{tabular}%
}

\small 
\caption{QA performance (\%) using nvidia/NVEmbed-v2 as the Retriever and Llama-3-8B-Instruct as Backbone. The best and second-best performances are highlighted in bold and underlined, respectively.}
\label{table:qa_performance_e5_different_readers}
\end{table}

\subsection{Effect of the Number of Candidate Triples} 
\label{app:effect_of_n}
During the construction phase of \textsc{CIRAG}, the reranker retrieves the top-$N$ triples most relevant to the current query, which are treated as the candidate triple set. To study the sensitivity to $N$, we vary $N$ from 10 to 100. Figure~\ref{figure:effect_num_candidate_triples} reports the QA performance under different $N$ values, over 3,000 questions sampled from three multi-hop QA datasets (the same as in the main experiment). Overall, \textsc{CIRAG} exhibits low sensitivity to $N$: both F1 and EM remain largely stable, with fluctuations below 2\% across the tested range. This robustness is consistent with our integration module, which refines the candidate set into a compact core triple set by leveraging the accumulated history, thereby mitigating the effect of noisy candidates. Considering cost and efficiency, we chose N = 30.
\begin{table}[t] 
\centering
\resizebox{\columnwidth}{!}{%
\begin{tabular}{lcccccc}
\toprule
\multirow{2}{*}{\textbf{Method}} & \multicolumn{2}{c}{\textbf{2WikiMQA}} & \multicolumn{2}{c}{\textbf{HotpotQA}} & \multicolumn{2}{c}{\textbf{MuSiQue}} \\
\cmidrule(lr){2-3} \cmidrule(lr){4-5} \cmidrule(lr){6-7}
 & \textbf{F1} & \textbf{EM} & \textbf{F1} & \textbf{EM} & \textbf{F1} & \textbf{EM} \\
\midrule
IRCOT & 44.8 & 35.9 & 55.1 & 41.2 & 21.8 & 12.2 \\
FLARE & 41.2 & 34.4 & 54.1 & 39.6 & 21.3 & 11.8 \\
MetaRAG & 48.6 & 42.8 & 60.3 & 47.6 & 29.2 & 18.8 \\
Kirag & 51.9 & 36.9 & 61.5 & \underline{48.1} & 30.7 & 19.5 \\
Dualrag & 61.2 & 51.1 & 57.6 & 44.5 & 32.9 & 21.0 \\
Dualrag-FT & \underline{64.5} & \underline{52.3} & \underline{61.7} & 46.7 & \underline{35.2} & \underline{24.8} \\
\midrule
\textbf{Ours} & \textbf{66.8} & \textbf{57.3} & \textbf{65.6} & \textbf{51.8} & \textbf{39.9} & \textbf{27.8} \\
\bottomrule
\end{tabular}%
}
\small 
\caption{QA performance (\%) using bge-Small-env1.5 as the Retriever and Qwen2.5-7B-Instruct as Backbone. The best and second-best performances are highlighted in bold and underlined, respectively.}
\label{table:retrieval_performance_bge}
\end{table}

\subsection{Effect of the number of iterative steps L} 
\label{app:effect_of_L}
Figure~\ref{figure:effect_L} reports the QA performance of \textsc{CIRAG} and the iRAG baseline on our 2WikiMQA test set under different maximum iteration steps $L$. As $L$ increases, the performance of methods generally improves in the first few steps and then plateaus. Notably, \textsc{CIRAG} CIRAG achieves optimal performance at any value of $L$ and already stabilizes at $L=2$, indicating that it can achieve strong accuracy with fewer rounds and thus improves efficiency.

\section{Case Study}
\label{appendix:case}

We conducted several case studies to analyze the effectiveness of our method. The results show that DualRAG can dynamically retrieve information when additional knowledge is required, enabling efficient multi-hop reasoning with minimal iterations. Detailed case study examples are illustrated in Table~\ref{tab:case}.

\begin{table*}[htbp]
\centering
\small
\setlength{\tabcolsep}{2.5pt}
\begin{tabular}{p{0.95\linewidth}}

\toprule

\multicolumn{1}{c}{\textbf{Case Study}} \\

\midrule

\textbf{Question}: Which film has the director who is older, God'S Gift To Women or Aldri Annet Enn Bråk? \\
\textbf{Gold Answers}: [ God'S Gift To Women ] \\
\midrule
\textbf{Iterative Construction-Integration (ICI) Retrieval module} \\
\midrule\\
\textbf{Step 1} \\

\textbf{Construction Phase} \\

\textcolor{brown}{Retrieved Documents} \\ 

(1) God's Gift to Women: El extraño viaje is a 1964 Spanish black drama film directed by Fernando Fernán Gómez... \\
(2) Aldri annet enn bråk: Aldri annet enn bråk is a 1954 Norwegian comedy- drama film... \\
(3) Dan Milne: Dan Milne is a British actor/ director who is possibly best known for his role... \\
(4) ... \\
\\

\textcolor{brown}{Candidate Triples} \\ 

( "god s gift to women", "directed by", "michael curtiz" ), ( "aldri annet enn br k", "directed by", "edith carlmar" ), ( "god s gift to women", "is a", "romantic musical comedy film" )...\\
\\

\textbf{Integration Phase} \\
\textcolor{brown}{Thought} \\ 
The query asks: Which film has the director who is older, God's Gift to Women or Aldri Annét Enn Bråk? From the facts provided, we know that God's Gift to Women was directed by Michael Curtiz and Aldri Annét Enn Bråk was directed by Edith Carlmar. To determine which director is older, we need their birth years. The next step is to find out the birth years of Michael Curtiz and Edith Carlmar.\\
\\

\textcolor{brown}{Core Triples} \\ 

( "god s gift to women", "directed by", "michael curtiz" ), ( "aldri annet enn br k", "directed by", "edith carlmar" )\\
\\

\textcolor{brown}{Next-hop query} \\ 
What are the birth years of Michael Curtiz and Edith Carlmar?\\
\\
\midrule
\textbf{Step 2} \\

\textbf{Construction Phase} \\

\textcolor{brown}{Retrieved Documents} \\ 

(1) Edith Carlmar: Edith Carlmar (Edith Mary Johanne Mathiesen) (15 November 1911 - 17 May 2003)... \\
(2) Michael Curtiz: Michael Curtiz (born Manó Kaminer(1886-1905) Mihály Kertész ... \\
(3) ... \\
\\

\textcolor{brown}{Candidate Triples} \\ 

( "edith carlmar", "started", "carlmar film a s" ), ( "edith carlmar", "is also known as", "edith mary johanne mathiesen" ), ( "edith carlmar", "born on", "15 november 1911" ), ( "michael curtiz", "born on", "december 24  1886" ),...\\
\\

\textbf{Integration Phase} \\
\textcolor{brown}{Thought} \\ 
The question is: Which film has the director who is older, God's Gift to Women or Aldri Annét Enn Bråk? From the previous step, we identified that Michael Curtiz directed God's Gift to Women and Edith Carlmar directed Aldri Annét Enn Bråk. The current facts provide their birth dates: Michael Curtiz was born on December 24, 1886 and Edith Carlmar was born on November 15, 1911. Since Michael Curtiz is older than Edith Carlmar. Therefore, the film God's Gift to Women has the older director.\\
\\

\textcolor{brown}{Core Triples} \\ 

( "edith carlmar", "born on", "15 november 1911" ), ( "michael curtiz", "born on", "december 24  1886" )\\
\\

\textcolor{brown}{Next-hop query} \\ 
<no question>\\
\\
\midrule
\textbf{Adaptive Cascaded Multi-Granularity Knowledge-Enhanced Generation (ACMG) module} \\
\midrule
\textcolor{brown}{Cumulative Triple Set} \\ 
( "god s gift to women", "directed by", "michael curtiz" ), ( "aldri annet enn br k", "directed by", "edith carlmar" ), ( "edith carlmar", "born on", "15 november 1911" ), ( "michael curtiz", "born on", "december 24  1886" )\\
\\

\textcolor{brown}{Thought} \\ 
Michael Curtiz (born Dec 24, 1886) is older than Edith Carlmar (born Nov 15, 1911). Therefore, the film with the older director is God'S Gift To Women.

Answer: God'S Gift To Women\\
\\
\bottomrule
\end{tabular}
\caption{Case Study}
\label{tab:case}
\end{table*}

\begin{figure*}[ht]
\centering
\begin{tcolorbox}[colback=gray!5!white,colframe=gray!60!black,title=Prompt Used for NER,left=0.2mm, right=0.2mm,bottom=0.2mm]
\begin{normalsize}
    \textbf{Instruction}: Your task is to extract named entities from the given paragraph. Respond with a JSON list of entities.
    
    \vspace{0.5em} \textbf{Examples}: 
    
    {\fontsize{10}{4}\selectfont {Paragraph}: \textit{Radio City
    Radio City is India's first private FM radio station and was started on 3 July 2001. It plays Hindi, English and regional songs. Radio City recently forayed into New Media in May 2008 with the launch of a music portal - PlanetRadiocity.com that offers music related news, videos, songs, and other music-related features.}}

    {\fontsize{10}{4}\selectfont {Entity lists}: \textit{\{"named entities":
    ["Radio City", "India", "3 July 2001", "Hindi", "English", "May 2008", "PlanetRadiocity.com"]
\}}}

    \vspace{0.5em} \textbf{Inputs}:

    {\fontsize{10}{4}\selectfont {Passage}: \textit{\{\textit{document text}\}}}
    
    \vspace{0.5em} \textbf{Outputs}:
    
    \fontsize{10}{4}\selectfont {Entity lists}:
    
\end{normalsize}
\end{tcolorbox}
\vspace{-1.0em}
\caption{Prompt used for NER.}
\label{figure:prompt_NER}
\vspace{-0.5em}
\end{figure*}

\begin{figure*}[ht]
\centering
\begin{tcolorbox}[colback=gray!5!white,colframe=gray!60!black,title=Prompt Used for Knowledge Triple Extraction,left=0.2mm, right=0.2mm,bottom=0.2mm]
\begin{normalsize}
    \textbf{Instruction}: Your task is to construct an RDF (Resource Description Framework) graph from the given passages and named entity lists. Respond with a JSON list of triples, with each triple representing a relationship in the RDF graph. 
    Pay attention to the following requirements:
    - Each triple should contain at least one, but preferably two, of the named entities in the list for each passage.
    - Clearly resolve pronouns to their specific names to maintain clarity.
    
    \vspace{0.5em} \textbf{Examples}: 
    {\fontsize{10}{4}\selectfont {Paragraph}: \textit{Radio City
    Radio City is India's first private FM radio station and was started on 3 July 2001. It plays Hindi, English and regional songs. Radio City recently forayed into New Media in May 2008 with the launch of a music portal - PlanetRadiocity.com that offers music related news, videos, songs, and other music-related features.}}

    {\fontsize{10}{4}\selectfont {Entity lists}: \textit{\{"named entities":
    ["Radio City", "India", "3 July 2001", "Hindi", "English", "May 2008", "PlanetRadiocity.com"]
\}}}

    {\fontsize{10}{4}\selectfont {Knowledge Triples}: \textit{\{"triples": [
            ["Radio City", "located in", "India"],
            ["Radio City", "is", "private FM radio station"],
            ["Radio City", "started on", "3 July 2001"],
            ["Radio City", "plays songs in", "Hindi"],
            ["Radio City", "plays songs in", "English"],
            ["Radio City", "forayed into", "New Media"],
            ["Radio City", "launched", "PlanetRadiocity.com"],
            ["PlanetRadiocity.com", "launched in", "May 2008"],
            ["PlanetRadiocity.com", "is", "music portal"],
            ["PlanetRadiocity.com", "offers", "news"],
            ["PlanetRadiocity.com", "offers", "videos"],
            ["PlanetRadiocity.com", "offers", "songs"]
    ]
\}}}

    \vspace{0.5em} \textbf{Inputs}:

    {\fontsize{10}{4}\selectfont {Title}: \textit{\{\textit{document title}\}}}
    
    {\fontsize{10}{4}\selectfont {Text}: \textit{\{\textit{document text}\}}}

    \vspace{0.5em} \textbf{Outputs}:

    {\fontsize{10}{4}\selectfont {Knowledge Triples}:}
    
\end{normalsize}
\end{tcolorbox}
\vspace{-1.0em}
\caption{Prompt used for extracting knowledge triples.}
\label{figure:prompt_knowledge_triple_extraction}
\vspace{-0.5em}
\end{figure*}

\begin{figure*}[ht]
\centering
\begin{tcolorbox}[colback=gray!5!white,colframe=gray!60!black,title=Prompt Used for Integration Phase,left=0.2mm, right=0.2mm,bottom=0.2mm]
\begin{normalsize}
    \textbf{Instruction}:  Your task is to loop through collecting facts to solve the query until enough facts are collected. You must plan to continue in a series of steps, looping as follows:[[ \#\# fact\_before\_filter \#\# ]]\{fact\_before\_filter\}[[ \#\# thought \#\# ]]\{thought\}[[ \#\# fact\_after\_filter \#\# ]]\{fact\_after\_filter\} [[ \#\# question \#\# ]] \{question\} sequence. In each step, in the [[ \#\# fact\_before\_filter \#\# ]] section, this is the collection of facts that need to be filtered to solve the previous problem. In the [[ \#\# thought \#\# ]] section, only use the information from the facts before filtering, briefly analyze the contribution of the facts to the problem, and list useful facts. Based on the analysis, determine whether the original problem can be solved. If it cannot be solved, analyze what additional information is needed and raise new questions. In the [[ \#\# fact\_after\_filter \#\# ]] section,  if there are no relevant facts, return an empty list:\{ "fact": [] \}.In the [[ \#\# question \#\# ]] section, you need to use the filtered facts to propose the next problem that still needs to be solved. If the facts collected in the previous rounds are sufficient to answer the original query, output \texttt{<no question>} as the end to terminate the loop.

    \vspace{0.5em} \textbf{i-th Inputs}:

    {\fontsize{10}{4}\selectfont {[[ \#\# Original Query \#\# ]]}: \textit{\{\textit{query}\}}}

    {\fontsize{10}{4}\selectfont {[[ \#\# fact\_before\_filter \#\# ]]}: \textit{\{\textit{fact\_before\_filter}\}}}

    {\fontsize{10}{4}\selectfont {[[ \#\# historical\_context \#\# ]]}: \textit{\{\textit{historical context}\}}}

    \vspace{0.5em} \textbf{i-th Outputs}:

    {\fontsize{10}{4}\selectfont {[[ \#\# thought \#\# ]]}:}
    
    {\fontsize{10}{4}\selectfont {[[ \#\# fact\_after\_filter \#\# ]]}:}

    {\fontsize{10}{4}\selectfont {[[ \#\# question \#\# ]]}:}
    
\end{normalsize}
\end{tcolorbox}
\vspace{-1.0em}
\caption{Prompt used for Integration Phase.}
\label{figure:prompt_Integration_Phase}
\vspace{-0.5em}
\end{figure*}

\begin{figure*}[ht]
\centering
\begin{tcolorbox}[colback=gray!5!white,colframe=gray!60!black,title=Prompt Used for triple level in ACMG.,left=0.2mm, right=0.2mm,bottom=0.2mm]
\begin{normalsize}
    \textbf{Instruction}: As an advanced reading comprehension assistant, your task is to carefully analyze triples and extract useful information from them to answer corresponding questions. Your response start after "Thought: ", where you will methodically break down the reasoning process, illustrating how you arrive at conclusions. If the provided information cannot answer the question, output Unanswerable. Conclude with "Answer: " to present a concise, definitive response, devoid of additional elaborations.
    
    \vspace{0.5em} \textbf{Examples}: 
    {\fontsize{10}{4}\selectfont {Triples}: \textit{triples: ('erik hort', 'born in', 'montebello'), ('erik hort', 'is', 'american'), ('montebello', 'is located in', 'rockland county')}}

    {\fontsize{10}{4}\selectfont {Query}: \textit{What county is Erik Hort's birthplace a part of?}}

    {\fontsize{10}{4}\selectfont {Thought}: \textit{The triplet ('erik hort', 'born in', 'montebello') tells us that Erik Hort was born in Montebello. The triplet ('montebello', 'is located in', 'rockland county') tells us that Montebello is located in Rockland County. Therefore, since Erik Hort was born in Montebello and Montebello is in Rockland County, Erik Hort's birthplace is part of Rockland County.}}

    {\fontsize{10}{4}\selectfont {Answer}: \textit{Rockland County.}}

    \vspace{0.5em} \textbf{Inputs}:

    {\fontsize{10}{4}\selectfont {Triples}: \textit{\{\textit{triples}\}}}
    
    {\fontsize{10}{4}\selectfont {Query}: \textit{\{\textit{query}\}}}

    \vspace{0.5em} \textbf{Outputs}:

    {\fontsize{10}{4}\selectfont {Thought}:}

\end{normalsize}
\end{tcolorbox}
\vspace{-1.0em}
\caption{Prompt used for triple level in ACMG.}
\label{figure:prompt_ACMG}
\vspace{-0.5em}
\end{figure*}

\end{document}